%% file: report_revised.tex
\documentclass[11pt]{article}
\usepackage{latexsym,epic,epsfig,amssymb,cite,epstopdf,multirow,rotating,epsfig,psfrag, art}
\usepackage{graphics,psfrag,amsmath,soul} 
\usepackage{pgfgantt}

\newcounter{mycounter}

\begin{document}

\pagenumbering{roman}

\input{title.tex}
\clearpage\baselineskip 5pt\parskip 4.85pt 
\clearpage \setcounter{page}{1}
\parskip 4pt
\baselineskip 13.7pt
\pagenumbering{arabic}

\section{Introduction}
Gait analysis of patients with neurological disorders, including multiple sclerosis (MS), is important for rehabilitation and treatment. In a clinical setting, gait analysis performed by physicians or therapists involves observing a patient's gait and manual documentation of subjective assessments. Different clinical scores are proposed in order to quantify the level of progression of MS. Among them the \emph{expanded disability status scale} (EDSS) and the self-scored \emph{multiple sclerosis walking scale} (MSWS) are widely used in clinical practice~\cite{Hobart, Kurtzke}. However, these clinical scores are subjective and are unable to detect subtle changes in a subject's gait due to the progression of the disease or the response to treatment. A more systematic gait analysis can be carried out in a gait laboratory using motion capture systems, force-plates, and electromyography (EMG) sensors. However, the required set up involves elaborate preparation and marker placement, rendering it unsuitable for use as a point-of-care technology.

A number of indices based on time or distance characteristics of human gait cycle have been proposed for objective gait assessment in neurological patients with gait abnormality. Specifically, studies have shown that a shorter stride length and gait swing time and a higher \emph{double support percentage} in a gait cycle (i.e., the fraction of the time in a gait cycle where the two feet are on the ground) are observed in neurological patients with gait abnormality~\cite{Holden, Remelius, Sacco, Givon}.
In addition, the range of the hip and knee angle is smaller in MS patients as compared to healthy individuals as reported in ~\cite{Gehlsen, Benedetti}. Correlations between time-distance and joint angle indices and the EDSS score have been investigated as well~\cite{Sacco, Huisinga, Givon}.

Angles of the lower extremity joints in a MS patient's gait have also been investigated. However, researchers have reported contradictory results. Specifically, reduced hip and knee flexion and ankle plantarflexion at \emph{heel strike} (i.e., the point in the gait cycle when foot reaches the ground), and hip and knee extension at \emph{toe-off} (i.e., the point in the gait cycle where foot is no longer in contact with the ground) are reported in~\cite{ Huisinga, Kelleher}, while increased hip and knee flexion and ankle plantarflexion at heel strike has been observed in~\cite{Benedetti}. The analysis in these studies involve point-to-point comparisons between preselected peak points of the joint angles during a gait cycle. 


Although accurate gait analysis for MS patients could provide valuable insight on a patient's condition and the progression of the disease, an easy-to-use system which can encourage clinical adoption does not exist. Motion capture systems, which are mainly used for experimental data collection for gait analysis, are expensive. Furthermore, motion capture systems can only track a set of reflective markers mounted on certain anatomical landmarks of a patient's body, and hence, its use requires a time-consuming patient preparation process. Moreover, operator training for motion capture systems is necessary to ensure acceptable accuracy. Finally, these systems are not portable and require a dedicated space which makes them further impractical for clinical applications. 

The Mircrosoft Kinect\textsuperscript{\textregistered} sensor, which was developed for motion recognition in gaming applications, is an ideal candidate for an inexpensive system providing the capability for human gait analysis. The Kinect sensor includes a color camera and a depth sensor, consisting of an infrared projector and camera, and provides full-body 3D motion capture. 
The Kinect sensor has been used for various clinical and non-clinical applications. The authors in~\cite{Shotton} use the Kinect sensor for pose identification. In \cite{Baena}, joint angles identified by the Kinect sensor are compared to the ``gold standard''  obtained by a marker-based motion capture systems for healthy subjects, which showed reasonable accuracy for clinical applications. The validity of the Kinect sensor for the assessment of the postural control was examined by comparing the result with the result of a marker-based motion capture system too~\cite{Clark}.  Also, the accuracy of this sensor for movement measurement in people with neurological disease, such as Parkinson, has been examined~\cite{Galna}. 

However, a limited number of studies have been performed to investigate the feasibility of Kinect specifically for gait analysis of MS patients~\cite{Pfueller, Souza, Behrens2}. In \cite{Behrens2}, the short maximum speed walk test was proposed by the authors to be measured with Kinect, where the correlation of this index with EDSS was investigated. Furthermore, based on machine learning and image processing techniques, movements of MS patients were compared with healthy subjects to identify subgroups with similar movement patterns \cite{Souza}. Authors in \cite{Kontschieder} have developed a framework to identify MS patients from healthy subjects by analyzing a number of tasks such as finger-to-nose and finger-to-finger tests. 

In this research, we develop a framework to quantify the gait abnormality of MS patients using a Kinect for Windows (version 1) camera. We show that previously introduced indices quantifying gait abnormality as well as an index for quantifying a patient's gait pattern introduced in this report are in strong agreement with clinical scoring systems that quantify the degree of the progression of the disease. Specifically, we show that there is a correlation between a number of gait indices and the self-scored MSWS as well as the \emph{clinical ambulation score} ~\cite{Hobart, Kurtzke} when clinical data is obtained using a Kinect camera. The clinical ambulation score, which is one of the required scores for computing the EDSS score, is determined by a clinician through clinical observation of a patient's gait. Considering the clinical ambulation score appears to be more meaningful for patient gait abnormality assessment compared to EDSS which quantifies a patient's general disability due to the MS disease and involves factors other than gait. 

In this report, in addition to the previously introduced gait indices, a novel set of MS gait indices based on the concept of \textit{dynamic time warping} \cite{Muller} is introduced. The newly introduced indices can characterize a patient's gait pattern as a whole (rather than considering isolated events in a gait cycle), and quantify a subject's gait ``distance'' from the healthy population. We will investigate the correlation of  these novel indices with the MSWS and the clinical ambulation score. This work establishes the feasibility of using the Kinect sensor for clinical gait assessment for MS patients.

\section{Gait Analysis for MS Patients}
\subsection{Subjects}

In this study, 10 male and female MS patients (9 females and 1 male), and 10 sex and age matched healthy control subjects were asked to walk in front of the Kinect camera for 5-10 \emph{trials} (i.e., a video sequence which involves a subject moving in front of the camera on a straight line). The best five captures were selected for the analysis. This study was approved by the Research Ethics Board at the McGill University Health Center and was conducted at the Montreal Neurological Institute. For each patient the MSWS and the clinical ambulation score were assigned by an on-site physician on the day of the study. MS patient information and their clinical assessment scores are summarized in Table~\ref{patient}, whereas Table~\ref{control} summarizes the information related to healthy control subjects. 

\begin{table}[h!]
\caption{{\small MS Subjects Specifications}}
\vspace{3 mm} 
\centering 
\begin{tabular}{c c c c c c c}
\hline\hline
Patient No. & Sex & Age & Ambulation Score & MSWS & Height[cm] & Weight[kg] \\[0.5ex] 
\hline 
\vspace{0.5 mm} {\small P1}  & \small F  & \small 53  & 6 & 89.6 &  {\small 146 } & {\small 55 } \\
\vspace{0.5 mm} {\small P2}  & \small F  & \small 41 & 0 & 25 & {\small 170 } & {\small 86 }  \\
\vspace{0.5 mm} {\small P3}  & \small M  & \small 79 & 5 & 47.9 & {\small 169 } & {\small 91 }\\
\vspace{0.5 mm} {\small P4}  & \small F  & \small 69 & 1 & 60.4 & {\small 150 } & {\small 81 } \\
\vspace{0.5 mm} {\small P5}  &\small F  & \small 75 & 5  & 72.9 & {\small 157 } & {\small 89 }  \\
\vspace{0.5 mm} {\small P6}  & \small F  &\small 60 & 6 & 81  &{\small 170 } & {\small 79 }  \\
\vspace{0.5 mm} {\small P7}  & \small F  & \small 55 & 4 & 83.3 & {\small 168 } & {\small 81 }\\
\vspace{0.5 mm} {\small P8}  & \small F  & \small 70 & 9 & 97.9 & {\small 178 } & {\small 81 } \\
\vspace{0.5 mm} {\small P9}  & \small F  & \small 53 & 6 & 75 & {\small 174 } & {\small 59 } \\
\vspace{0.5 mm} {\small P10} & \small F  & \small 55 & 1 & 45.8 & {\small 161 } & {\small 51 } \\
\end{tabular}
\label{patient}
\end{table}  

\begin{table}[h!]
\caption{{\small Healthy Control Subjects Specifications}}
\vspace{3 mm} 
\centering 
\begin{tabular}{c c c c c}
\hline\hline
Control Subject No. &  Sex & Age  & Height[cm] & Weight[kg] \\[0.5ex] 
\hline 
\vspace{0.5 mm} {\small C1} & {\small F } & {\small 53 } & {\small 160 } & {\small 81} \\
\vspace{0.5 mm} {\small C2} & {\small F } & {\small 36 } & {\small 165 } & {\small 70} \\
\vspace{0.5 mm} {\small C3} & {\small M } & {\small 80 } & {\small 185 } & {\small 135} \\
\vspace{0.5 mm} {\small C4} & {\small F } & {\small 71 } & {\small 163 } & {\small 82} \\
\vspace{0.5 mm} {\small C5} & {\small F } & {\small 72 } & {\small 159 } & {\small 50} \\
\vspace{0.5 mm} {\small C6} & {\small F } & {\small 62 } & {\small 147 } & {\small 59} \\
\vspace{0.5 mm} {\small C7} & {\small F } & {\small 57 } & {\small 170 } & {\small 63} \\
\vspace{0.5 mm} {\small C8} & {\small F } & {\small 67 } & {\small 170 } & {\small 63} \\
\vspace{0.5 mm} {\small C9} & {\small F } & {\small 50 } & {\small 168 } & {\small 69} \\
\vspace{0.5 mm} {\small C10} & {\small F } & {\small 51 } & {\small 165 } & {\small 53} \\
\end{tabular}
\label{control}
\end{table}  

In each trial, subjects were asked to walk at a normal pace. They were instructed to start their gait outside of the camera's field of view in order to ensure that their normal walking patterns have been established once they reach the capture zone. As the Kinect field of view is limited, depending on the stride length of the subjects, one or multiple gait cycles might have been captured. The stored information related to each subject was de-identified to ensure patient privacy. Depending on the level of progression of the disease, patients may have walked with or without the use of assistive devices. Normal subjects passed an interview with the on-site physician to ensure they did not suffer from any gait abnormalities and do not possess any condition that could affect their gait.

\subsection{Data Analysis}\label{sec:data_analysis}

\begin{figure}[h!]
\begin{center}
\epsfig{figure=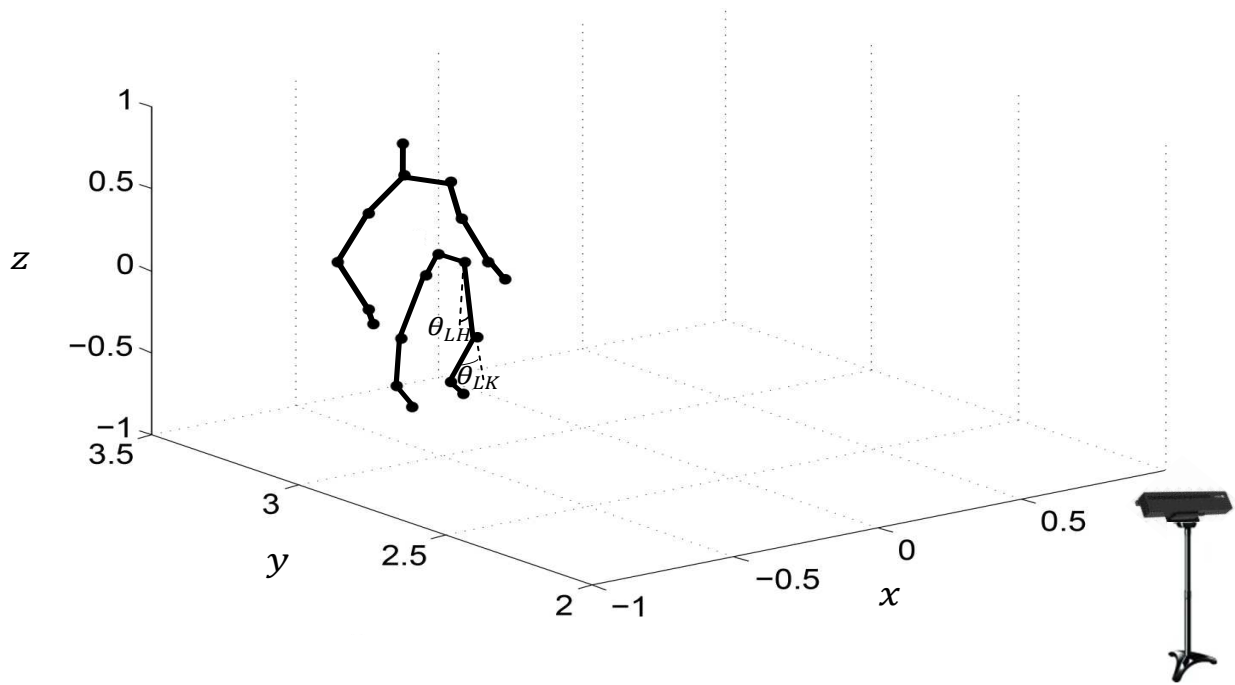, height=9cm, width=17cm}
\caption{\small{Captured data and the identified joints using Microsoft Kinect. The left hip angle $\boldsymbol{\theta}_\mathrm{LH}$ and left knee angle $\boldsymbol{\theta}_\mathrm{LK}$ are shown in the Figure as well.}} \label{gait}
\end{center}
\end{figure}

The Kinect for Windows sensor with the use of its software development kit (SDK) provides three-dimensional skeletal data on 20 joint positions over time. For the lower extremity, these joints consist of pelvis, knee, and ankle for each leg. Kinect captures the video up to 30 frames per second. Joint positions are expressed in an inertial reference frame in which the $y$-axis is in the direction of the runway, the $z$-axis is perpendicular to the ground, and the $x$-axis is mutually perpendicular to both (see Figure~\ref{gait}). 

\begin{figure}[h!]
  \begin{minipage}[b]{0.5\linewidth}    \centering
   \psfrag{T_HS}{\small{$T_{HS}$}}
   \psfrag{T_TO}{\small{$T_{TO}$}}
   \psfrag{T_TS}{\small{$T_{TS}$}}
    \includegraphics[width=\linewidth]{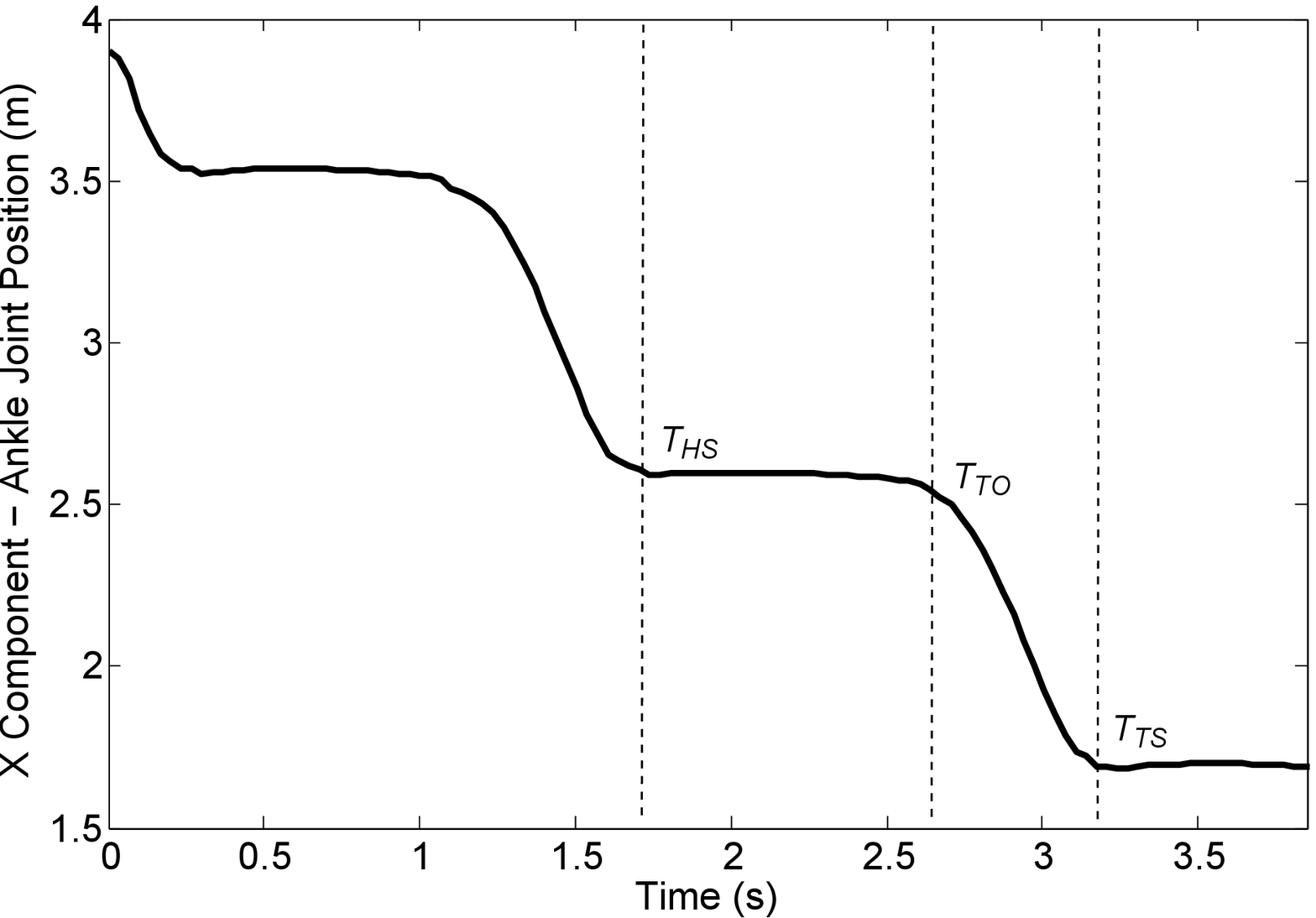}
    \caption{{Ankle joint position variation captured for Patient 6 (Trial 5). Heel strike, toe-off and terminal swing phases are denoted by $T_{\rm HS}$, $T_{\rm TO}$, and $T_{\rm TS}$, respectively.}}
    \label{ankle}
  \end{minipage}
  \hspace{1cm}
  \begin{minipage}[b]{0.5\linewidth}
    \centering
    \includegraphics[width=\linewidth]{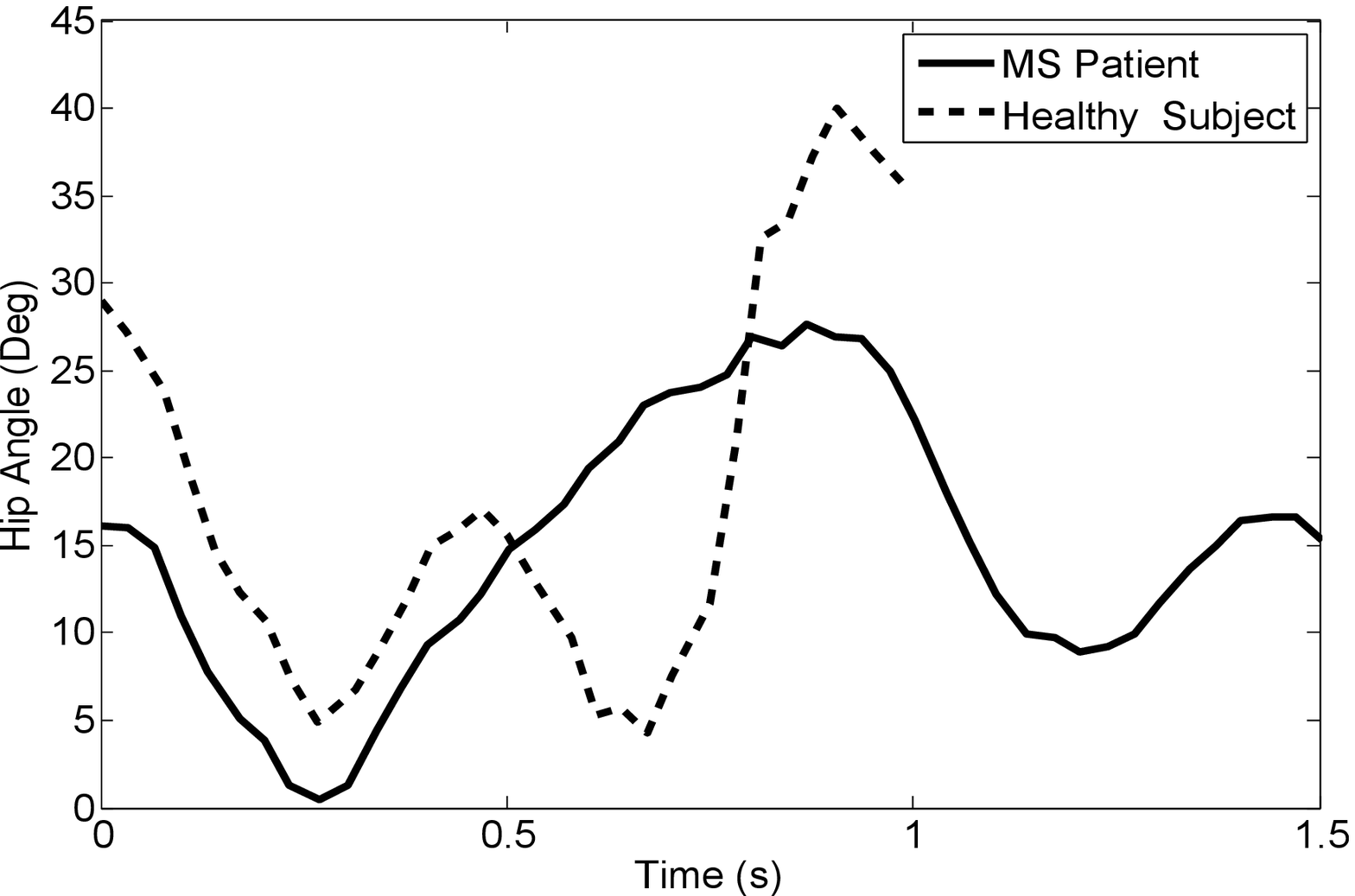}
    \caption{{Hip angle variations for an MS patient and a healthy control subject.}}
    \label{Hip}
  \end{minipage}
\end{figure}

The subject's kinematic properties can be extracted using the value of the joint positions. Joint angles, and if required, angular velocities and accelerations can be calculated based on the time-history of the joint positions. The first step in the process involves identifying a complete gait cycle in the captured data. This can be accomplished by considering ankle position variations over time. A gait cycle starts with heel strike, implying that the ankle joint position is stationary (see $T_\mathrm{HS}$ in Figure~\ref{ankle}). At toe-off the leg starts to swing, and hence, the ankle position starts to change (see $T_\mathrm{TO}$ in Figure~\ref{ankle}). Finally, the gait cycle terminates by the \textit{terminal swing} in which the ankle position comes to rest again (see $T_\mathrm{TS}$ in Figure~\ref{ankle}). The ankle joint position variation for a representative subject is shown in Figure~\ref{ankle}. 

In order to distinguish an abnormal gait from normal, and potentially quantify its degree of abnormality, appropriate gait indices need to be defined. Three categories of indices are defined for gait analysis in this study; namely, time-distance indices, joint angle, which have been discussed in the literature, and gait pattern, which is introduced in this report. Here, four time-distance indices, namely, subject velocity, stride length, stance percentage, and stride width are considered as suggested by authors in~\cite{Holden, Remelius, Sacco}. In the joint angle category, hip and knee range of motion is considered due to the fact that studies show that MS can substantially affect these joints~\cite{Gehlsen}. Finally, a novel index, which can capture the general gait pattern of the subject and its deviation from a healthy gait, is introduced.   

A subject's velocity for one gait cycle can be calculated by dividing the stride length to the gait cycle time (from $T_\mathrm{HS}$ to $T_\mathrm{TS}$). To be able to meaningfully compare distance indices, they are normalized based on a subject's height to compute \textit{normalized velocity} $\mathbf{V}_\mathrm{n}$. Furthermore,  \textit{normalized stride length} $\mathbf{L}_\mathrm{n}$, defined as the distance travelled by the ankle in one gait cycle (i.e., between the heel strike $T_\mathrm{HS}$ and the terminal swing phases $T_\mathrm{TS}$),  \textit{stride width} $\mathbf{W}$, defined as the distance between the two ankles in the double support phase projected on the $x$-axis, and \textit{stance percentage} $\mathbf{S}$, defined as the ratio of the stance time to the gait cycle time, can be calculated. Finally, the knee and hip angles need to be computed using the location of the joints. Knee angle is defined as the angle between the thigh and the leg segments, and the hip angle is defined as the angle between the $z$-axis and the thigh segment  (see Figure~\ref{gait}).

\subsubsection{Dynamic Time Warping}

Although minimum/maximum joint angle values, range of motion, and time-distance indices can provide valuable information on a subject's gait characteristics, they only provide a snapshot at a specific time instant. Developing a framework capable of analyzing a complete gait cycle (as opposed to analyzing certain points in a gait cycle) provides a holistic approach for gait analysis and can complement the information provided by other indices. This involves developing a distance metric to quantify the level of abnormality of a gait cycle with respect to healthy individuals. There have been prior attempts in comparing MS gait patterns with control subjects by point-to-point comparisons at certain gait phases such as heel strike or toe-off~\cite{Huisinga, Kelleher, Benedetti}. However, no study has introduced a mechanism to compare complete gait cycles. 

Here, we propose a set of novel MS gait indices based on the \textit{dynamic time warping} (DTW) framework. DTW, which was initially proposed for speech recognition applications~\cite{Rabiner}, provides a framework to find an optimal alignment between two time series that have different time scales. This is ideal for comparing sequences representing the human gait cycle as any gait cycle includes the same gait phases (i.e., heel strike, toe-off, etc.), however, the transition time between these phases varies from subject to subject. DTW has been previously used in other non-clinical contexts such as human motion recognition and in identifying different modes of movements and motion patterns~\cite{Veeraraghavan, Blackburn}. DTW defines a cost function and uses a nonlinear transformation to warp the two sequences in order to minimize the cost function. The optimal value of the cost function can be regarded as a ``distance measure'' between the two sequences.

Let us consider two sequences $\mathbf{A}= \{a_1, a_2, \dots, a_N\}$, $a_n\in\mathbb{R}$, $n=1,\,\dots,\,N$, and $\mathbf{B}=\{b_1, b_2, \dots,$ $b_M\}$, $b_m\in\mathbb{R}$, $m=1,\,\dots,\,M$. A \textit{local distance cost} between two elements of the sequence $a_n\in\mathbf{A}$ and $b_m\in\mathbf{B}$ is a mapping $c:A\times B\rightarrow \overline{\mathbb{R}}_+$ such that $c(a_n,b_m)$ increases as the mismatch between $a_n$ and $b_m$ increases. Next, we define a \emph{cost matrix} $\mathbf{C}=[c_{nm}] \in \mathbf{R}^{N \times M} $, where $c_{nm}=c(a_n,b_m)$, $n=1,\dots,N$, $m=1,\dots,M$. A sequence $p=\{p_1,\, p_2,\, \dots,\, p_L\}$, $p_l=(n_l,m_l)$, $l=1,\,\dots,\, L$, $n_l\in\{1,\,\dots,\, N\}$, $m_l\in\{1,\,\dots,\, M\}$ is referred to as a \emph{warping path}  if it satisfies the boundary, monotonicity, and step size conditions discussed in~\cite{Muller}. A warping path defines the point-to-point correspondence between the two sequences. The \textit{total cost} $c_p(\mathbf{A},\mathbf{B})$ for the two sequences $\mathbf{A}$ and $\mathbf{B}$ along a warping path $p$ is defined as
\begin{equation}
c_p(\mathbf{A},\mathbf{B}):=\sum\limits_{l=1}^L c(a_{n_l}, b_{m_l}),\quad a_{n_l}, b_{m_l}\in p.
\end{equation}

Consider the set of all possible warping paths for the two sequences $\mathbf{A}$ and $\mathbf{B}$ denoted by $\mathbf{P}$. The \textit{dynamic time warping distance} between $\mathbf{A}$ and $\mathbf{B}$ is defined \begin{eqnarray}
\mathrm{DTW}(\mathbf{A},\mathbf{B}):=c_{p^*}(\mathbf{A},\mathbf{B}),\label{eq:defdtw}
\end{eqnarray}
where $p^* := {\rm argmin}_{p \in \mathbf{P}} c_p(\mathbf{A},\mathbf{B})$ is the \textit{optimal warping path}.

\subsubsection{Dynamic Time Warping for Gait Analysis}

Here, we investigate the application of DTW to quantify MS disease progression. Specifically, we investigate its correlations with the MSWS and clinical ambulation score. The DTW framework can be used to align two sequences, for instance unaligned hip angle variations over time for a patient and a normal subject gait. Figure~\ref{Hip} shows the the two sequences for a representative patient and a control subject. The value of the DTW distance defined in (\ref{eq:defdtw}) shows the deviation of a patient's joint pattern from a normal subject. A larger DTW indicates a larger deviation between patient and the healthy subject gait patterns, and potentially a more advanced stage of MS. In longitudinal studies, all of these novel DTW distances could potentially show the progression of the MS disease in a particular patient over time and provide insight on a patient's response to treatments.

In this research, in order to evaluate a patient's gait, we extract time series for hip and knee angles for a complete gait cycle and compute the hip and knee DTW distances with respect to a set of control subjects. Note that maximum hip and knee flexion angles are reported to be different in MS patients compared to healthy control subjects~\cite{Huisinga, Kelleher, Benedetti}. In this study, the goal is to further investigate the relationship between the degree of gait abnormality and lower extremity joint variations over a complete gait cycle and not limit the analysis to extreme joint angles.  

The dataset included in this study is composed of gait data collected from $n_{\rm p}$ patients, where each patient completed $m_{\rm p}$ trials. In addition, the control dataset is composed of gait data collected from $n_{\rm c}$ control subjects, where each control subject completed $m_{\rm c}$ trials. As discussed in Section~\ref{sec:data_analysis}, for each trial, one complete gait cycle from heel strike to toe-off is identified and the rest is discarded. Therefore, overall the dataset includes $n_{\rm p} m_{\rm p}$ gait cycles corresponding MS patients and $n_{\rm c}  m_{\rm c}$ gait cycles corresponding to healthy control subjects. In our study, $n_{\rm p}=n_{\rm c}=10$ and $m_{\rm p}=m_{\rm c}=5$. 

For each patient, hip joint angles for the left and right legs over the entire gait cycle are calculated and stored in arrays $\boldsymbol{\theta}_{\mathrm{{LH}}_{i,j}}$, $\boldsymbol{\theta}_{\mathrm{{RH}}_{i,j}}$, respectively, where $i \in \{1,\dots, n_{\rm p}\}$, and $j \in \{1,\dots,m_{\rm p}\}$. Similarly, knee joint angles for the patient's left and right legs are stored as $\boldsymbol{\theta}_{\mathrm{{LK}}_{i,j}}$, $\boldsymbol{\theta}_{\mathrm{{RK}}_{i,j}}$, respectively, where $i \in \{1,\dots, n_{\rm p}\}$ and $j \in \{1,\dots, m_{\rm p}\}$. For control subjects, hip and knee joint angles for the left and right legs are calculated and stored in $\boldsymbol{\phi}_{\mathrm{{LH}}_{q,r}}$, $\boldsymbol{\phi}_{\mathrm{{RH}}_{q,r}}$, $\boldsymbol{\phi}_{\mathrm{{LK}}_{q,r}}$, $\boldsymbol{\phi}_{\mathrm{{RK}}_{q,r}}$, respectively, where $q \in \{1,\dots, n_{\rm c}\}$ and $r \in \{1,\dots, m_{\rm c}\}$. 

Here, we introduce an index referred to as the \textit{mean dynamic time warping distance}. This index quantifies the degree of dissimilarity between a patient's joint angle pattern compared to a set of control subjects. Specifically, the DTW distance between two time series, namely, a patient's joint angle for a given trial and a set of control subjects joint angles for all available trials are computed. Note that these distances are computed for the left and right legs independently. That is, a patient's left (right) leg joint angle sequence is compared with all left (right) joint angle sequences for all control subjects. Hence, overall $n_{\rm c} m_{\rm c}$ DTW distances are computed for each leg (in our case $n_{\rm c} m_{\rm c}= 50$). 

Considering the fact that $m_{\rm p}$ trials are available for each patient, $m_{\rm p} n_{\rm c} m_{\rm c}=250$ DTW distances are obtained for each patient's leg and each joint. Hence, a total of $2m_{\rm p} n_{\rm c} m_{\rm p}=500$ DTW distances are computed for each joint (i.e, knee and hip). These distances are then averaged to compute the mean dynamic time warping distance associated with the knee or hip joint for each patient. Specifically, for Patient $i$, $i \in \{ 1, \dots, n_{\rm p} \}$, the mean dynamic time warping distance for knee and hip joints, denoted by $\mathbf{D}_{\mathrm{K}_{\rm P}}$ and $\mathbf{D}_{\mathrm{H}_{\rm P}}$, respectively, are defined as
\begin{eqnarray}
\mathbf{D}_{\mathrm{K}_{\rm P}}&:=& \frac{1}{2} \bigg[\frac{1}{m_{\rm p}  n_{\rm c}  m_{\rm c}}\sum \limits_{j=1}^{m_{\rm p}} \sum \limits_{q=1}^{n_{\rm c}} \sum \limits_{r=1}^{m_{\rm c}} {\rm DTW}(\boldsymbol{\theta}_{{\mathrm{LK}}_{i,j}},\boldsymbol{\phi}_{{\mathrm{LK}}_{q,r}})+ \frac{1}{m_{\rm p}  n_{\rm c}  m_{\rm c}} \sum \limits_{j=1}^{m_{\rm p}} \sum \limits_{q=1}^{n_{\rm c}} \sum \limits_{r=1}^{m_{\rm c}} {\rm DTW}(\boldsymbol{\theta}_{{\mathrm{RK}}_{i,j}},\boldsymbol{\phi}_{{\mathrm{RK}}_{q,r}}) \bigg], \nonumber\\\label{eq:dtw1}\\
\mathbf{D}_{\mathrm{H}_{\rm P}}&:=& \frac{1}{2} \bigg[\frac{1}{m_{\rm p} n_{\rm c} m_{\rm c}}\sum \limits_{j=1}^{m_{\rm p}} \sum \limits_{q=1}^{n_{\rm c}} \sum \limits_{r=1}^{m_{\rm c}} {\rm DTW}(\boldsymbol{\theta}_{{\mathrm{LH}}_{i,j}},\boldsymbol{\phi}_{{\mathrm{LH}}_{q,r}}) + \frac{1}{m_{\rm p} n_{\rm c} m_{\rm c}} \sum \limits_{j=1}^{m_{\rm p}} \sum \limits_{q=1}^{n_{\rm c}} \sum \limits_{r=1}^{m_{\rm c}} {\rm DTW}(\boldsymbol{\theta}_{{\mathrm{RH}}_{i,j}},\boldsymbol{\phi}_{{\mathrm{RH}}_{q,r}}) \bigg]. \nonumber\\
\label{eq:dtw2}
\end{eqnarray}
Similarly, the mean dynamic time warping distance associated with the knee and hip joints for control subjects denoted by $\mathbf{D}_{\mathrm{K}_{\rm C}}$ and $\mathbf{D}_{\mathrm{H}_{\rm C}}$, respectively,  can be defined, where the distance between joint sequences for a control subject is compared with all other control subjects. 

\section{Results and Discussion}

The aforementioned time-distance and joint angle indices discussed in previous literature, and the newly introduced mean DTW indices associated with quantifying gait abnormality are computed for all patients and control subjects. A list of all the computed indices is given in Table~\ref{tab:index}. 
The mean value for each index (averaged over 5 trials) for patients and control subjects are 
summarized in Tables~\ref{indicesP} and \ref{indicesN}, respectively. 

\begin{figure}[h!]
\begin{center}
  \begin{minipage}[b]{0.42\linewidth}    \centering
    \includegraphics[width=\linewidth]{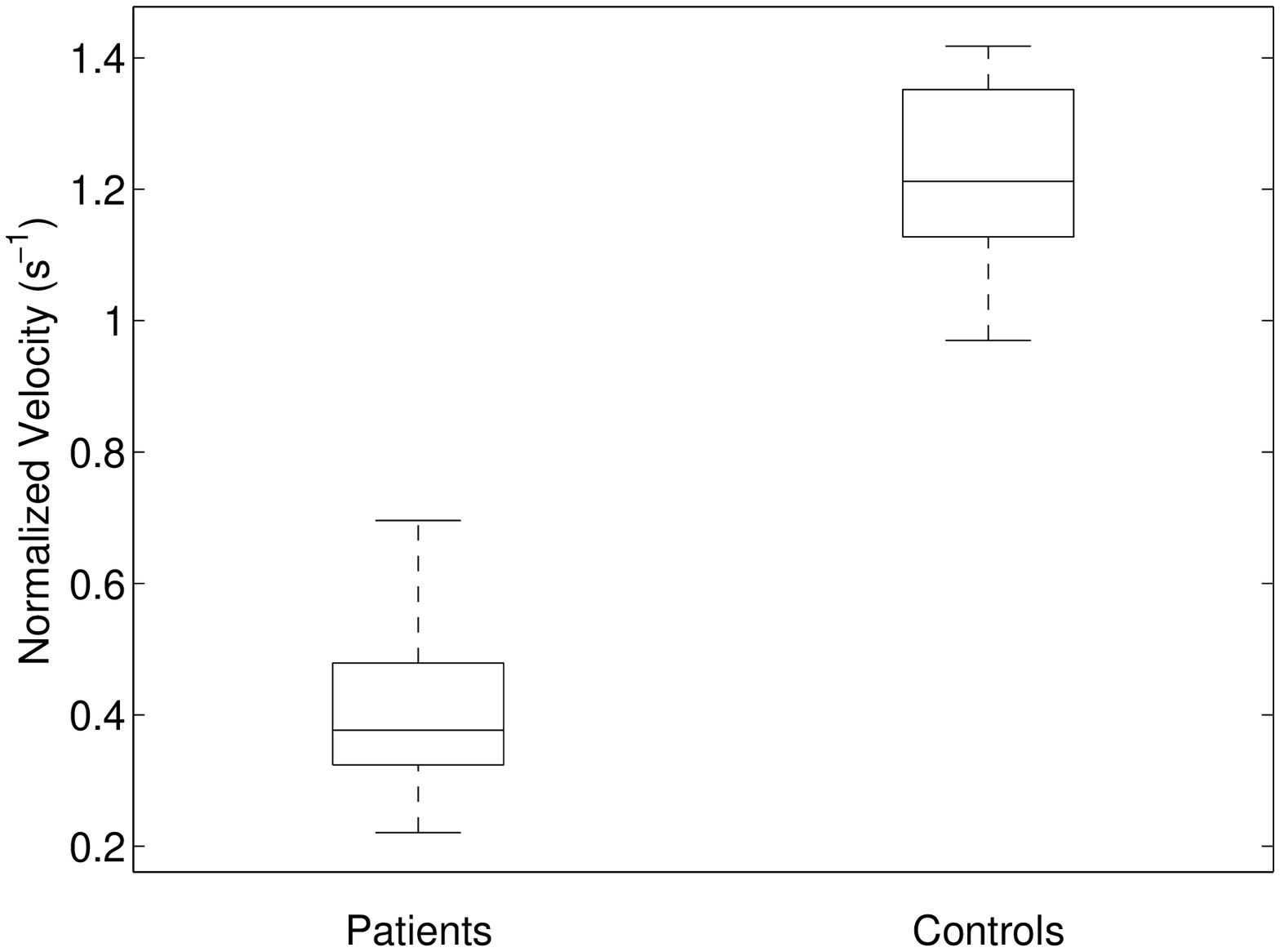}
    \caption{Normalized velocity $\mathbf{V}_{\rm n}$ for patients and control subjects.}
    \label{V}
  \end{minipage}
  \hspace{1cm}
  \begin{minipage}[b]{0.42\linewidth}
    \centering
    \includegraphics[width=\linewidth]{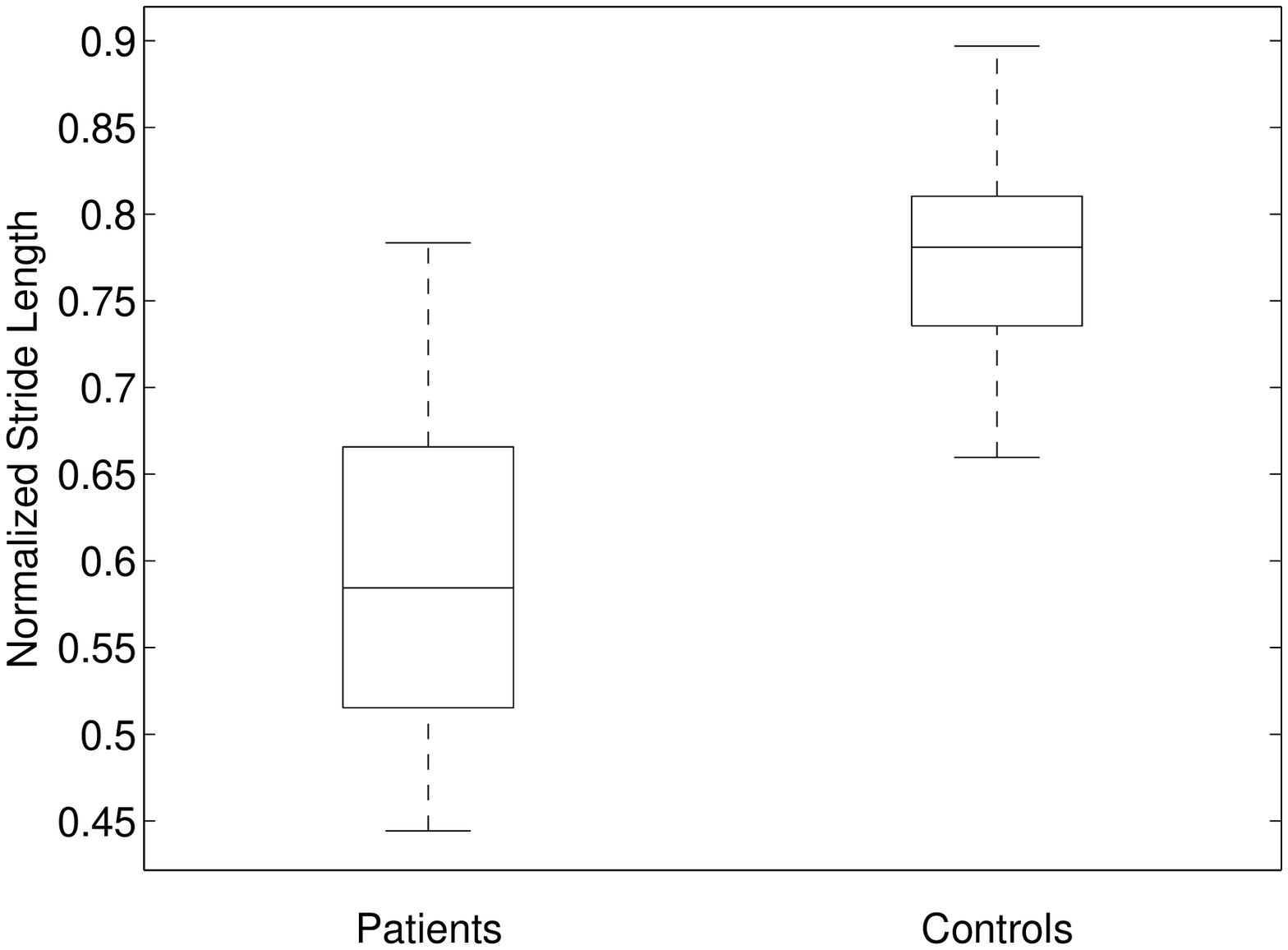}
    \caption{Normalized stride length $\mathbf{L}_{\rm n}$ for patients and control subjects.}
    \label{Stride}
  \end{minipage}
    \hspace{1cm}
    \end{center}
\end{figure}

\begin{figure}[h!]
\begin{center}
  \begin{minipage}[b]{0.42\linewidth}    \centering
    \includegraphics[width=\linewidth]{Box_Stance.eps}
    \caption{Stance percentage $\mathbf{S}$ for patients and control subjects.}
    \label{Stance}
  \end{minipage}
  \hspace{1cm}
  \begin{minipage}[b]{0.42\linewidth}
    \centering
    \includegraphics[width=\linewidth]{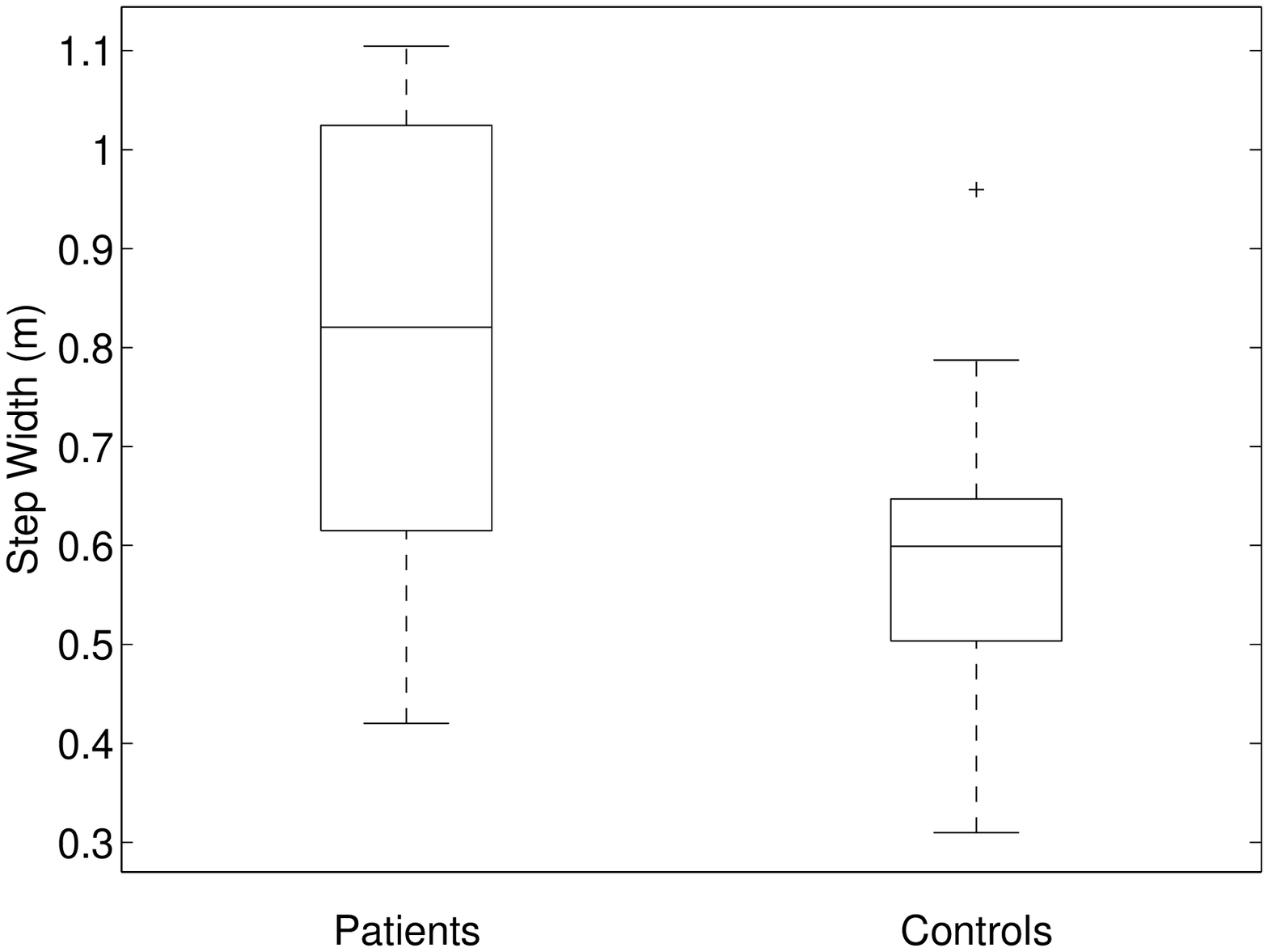}
    \caption{Step width length $\mathbf{W}$ for the patients and control subjects.}
    \label{KROM}
  \end{minipage}
    \hspace{1cm}
    \end{center}
\end{figure}

\begin{figure}[h!]
\begin{center}
  \begin{minipage}[b]{0.42\linewidth}    \centering
      \includegraphics[width=\linewidth]{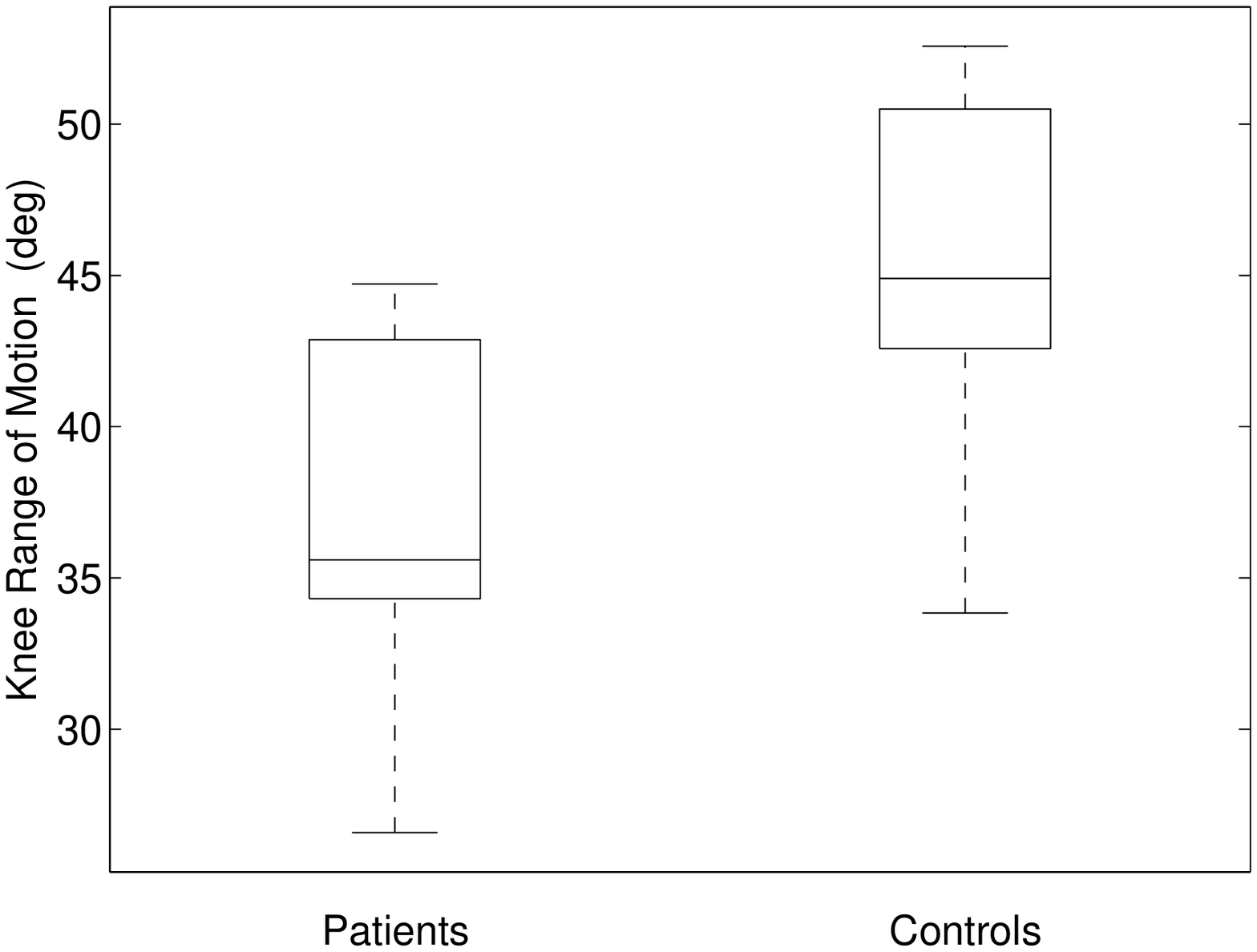}
    \caption{Knee range of motion $\boldsymbol{\alpha}_{\rm K}$) for patients and control subjects.}
    \label{KROM}  \end{minipage}
  \hspace{1cm}
  \begin{minipage}[b]{0.42\linewidth}
    \centering
    \includegraphics[width=\linewidth]{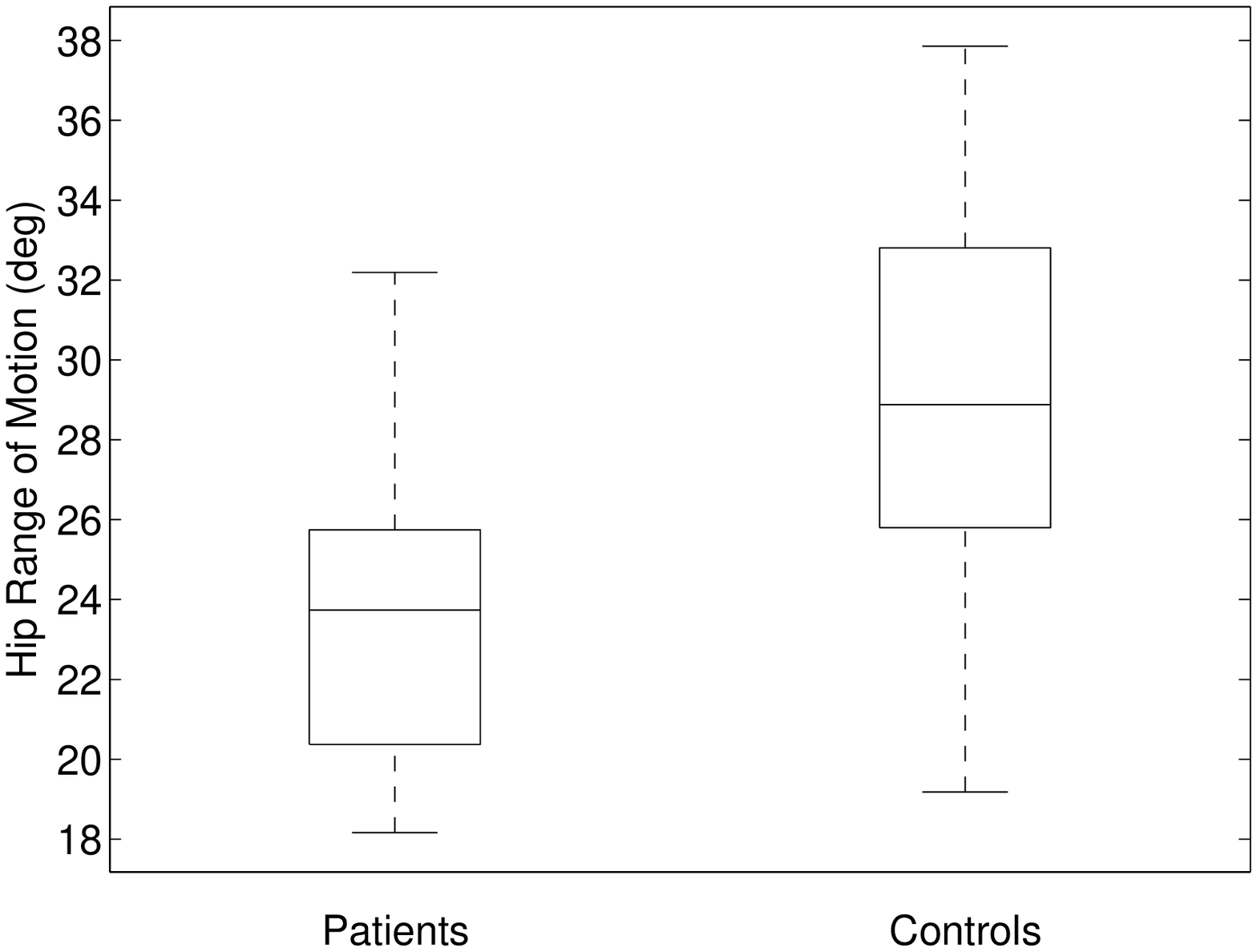}
    \caption {Hip range of motion $\boldsymbol{\alpha}_{\rm H}$) for patients and control subjects.}
    \label{HROM}
      \end{minipage}
    \hspace{1cm}
    \end{center}
\end{figure}

\begin{figure}[h!]
\begin{center}
  \begin{minipage}[b]{0.42\linewidth}    \centering
    \includegraphics[width=\linewidth]{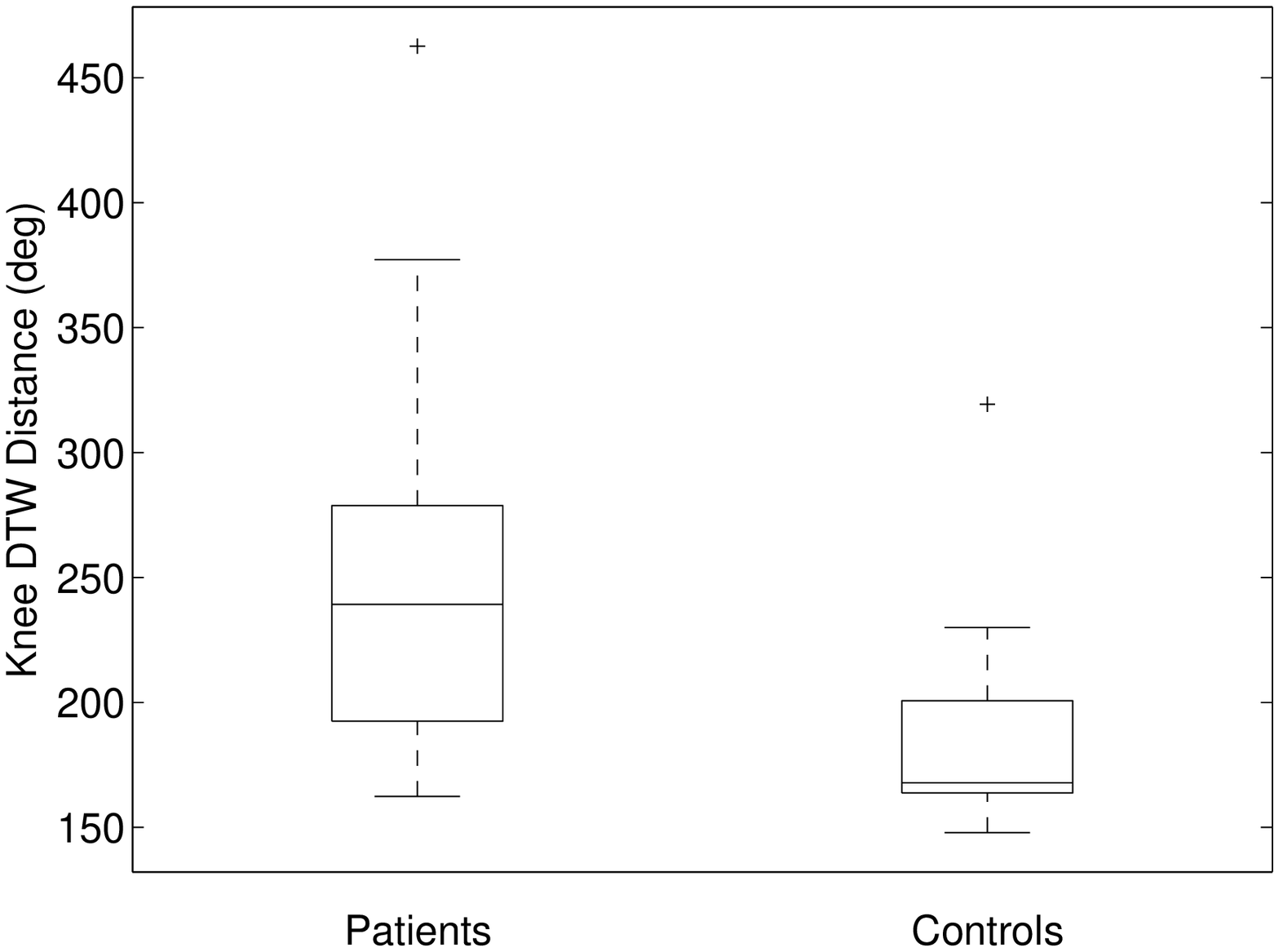}
    \caption{Knee mean DTW distance $\mathbf{D}_\mathrm{K}$ for patients and control subjects.}
    \label{KDTW}
     \end{minipage}
  \hspace{1cm}
  \begin{minipage}[b]{0.42\linewidth}
    \centering
    \includegraphics[width=\linewidth]{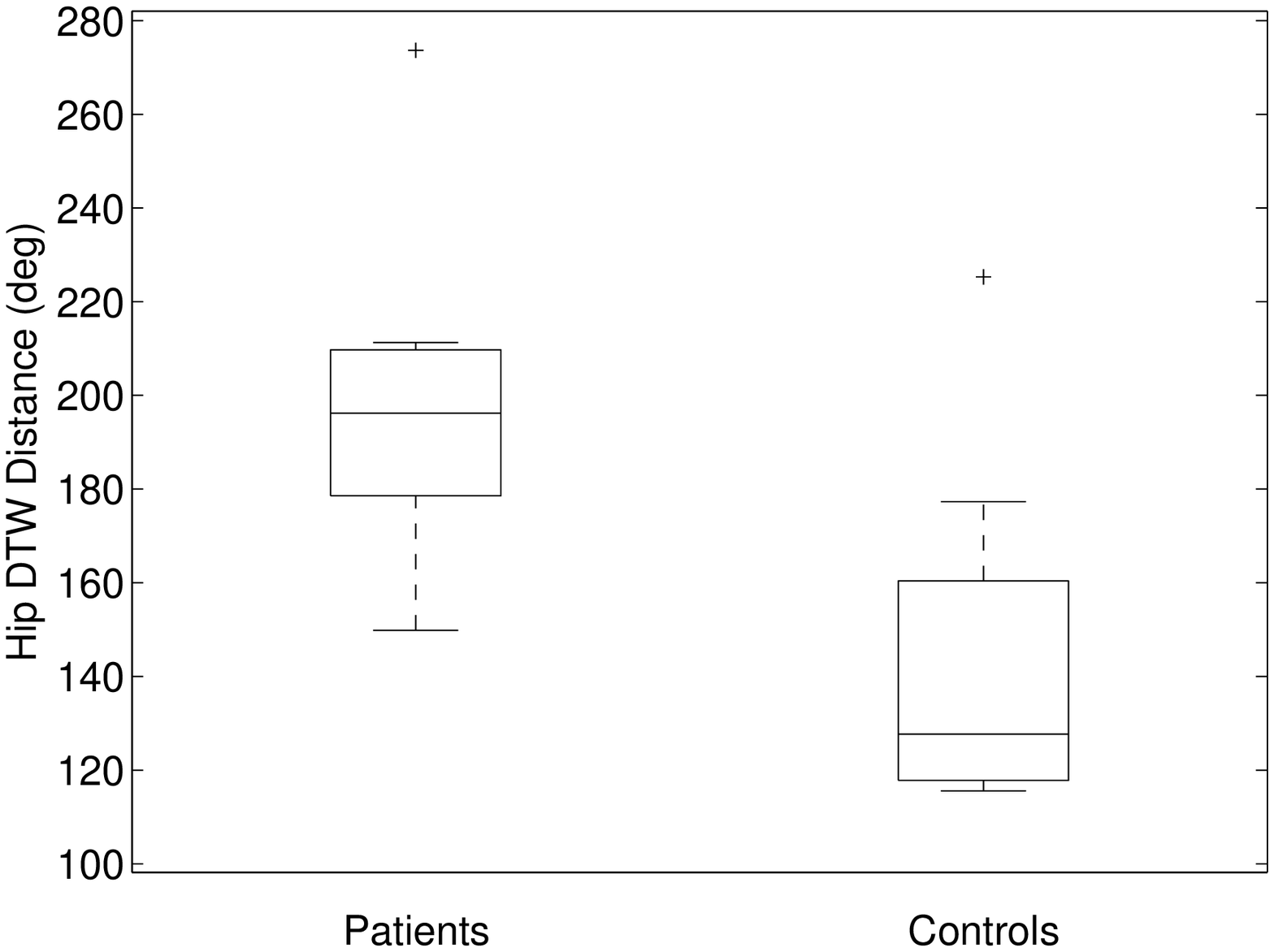}
    \caption{Hip mean DTW distance $\mathbf{D}_\mathrm{H}$ for patients and control subjects.}
    \label{HDTW}
    \end{minipage}
    \hspace{1cm}
    \end{center}
\end{figure}

The results are also illustrated in Figures~\ref{V}-\ref{HDTW} in form of box plots. The hip and knee positions for Patient 9 showed severe artifacts, and hence, this patient was excluded from the analysis for indices involving hip and knee. Note that for MS patients, the median velocity and median stride length are smaller compared to healthy control subjects (Figures~\ref{V} and \ref{Stride}), while the median stance percentage, as well as the knee and hip ranges of motions are larger (Figures~\ref{Stance}-\ref{HROM}). These observations are in agreement with previously reported gait characteristics of MS patients~\cite{Holden, Remelius, Sacco, Givon, Gehlsen, Benedetti}. Furthermore, the newly introduced indices based on the DTW distance can provide further insight on the general gait pattern. More specifically, the median hip and knee mean DTW distances increase significantly in subjects with MS disease (Figures~\ref{KDTW} and Figure~\ref{HDTW}). Larger DTW distances imply that the patient's gait pattern in the MS population is not ``similar'' to the healthy control group. 

\begin{table}[h!]
\caption{{\small Gait Indices for MS Patients}}
\vspace{3 mm} 
\centering 
\resizebox{0.9\textwidth}{!}{  
\begin{tabular}{l l l l}
\hline\hline
Index & Symbol & Unit & Description\\
\hline
Normalized velocity & $\mathbf{V}_\mathrm{n}$ & sec$^{-1}$ & Velocity normalized by hight\\
Normalized stride length & $\mathbf{L}_\mathrm{n}$ & Unitless & Stride length normalized by hight\\
Stance percentage & $\mathbf{S}$ & Unitless &  Stance time divided by gait cycle time\\
Step width & $\mathbf{W}$ & m & Distance between the two ankles in double support phase projected on the $x$-axis\\
Hip range of motion & $\boldsymbol{\alpha}_{\rm H}$ & deg & Difference between minimum and maximum of the hip angle\\
Knee range of motion & $\boldsymbol{\alpha}_{\rm K}$ & deg & Difference between minimum and maximum of the knee angle\\
Mean dynamic time warping distance for knee & $\mathbf{D}_\mathrm{K}$ & deg & See (\ref{eq:dtw1})\\
Mean dynamic time warping distance for hip & $\mathbf{D}_\mathrm{H}$ & deg & See (\ref{eq:dtw2})\\
\end{tabular}}
\label{tab:index}
\end{table}  

\begin{table}[h!]
\caption{{\small Gait Indices for MS Patients}}
\vspace{3 mm} 
\centering 
\resizebox{0.9\textwidth}{!}{  
\begin{tabular}{c c c c c c c c c c c}
\hline\hline
Patient & $\mathbf{V}_n$[s$^{-1}$] & $\mathbf{L}_n$ & $\mathbf{S}$[\%] & $\mathbf{W}$[m] &  $\boldsymbol{\alpha}_K$[deg] & $\boldsymbol{\alpha}_H$[deg] & $\mathbf{D}_\mathrm{K}$[deg]& $\mathbf{D}_\mathrm{H}$[deg]   \\[0.5ex] 
\hline 
\vspace{0.5 mm} {\small P1} & {\small 0.48 } & {\small 0.44 } & {\small 66 } &{\small 0.68} & {\small 36.0}  & {\small 17.9}  & {\small 191} &{\small 182}  \\
\vspace{0.5 mm} {\small P2} & {\small 1.23} & {\small 0.78} & {\small 59} &{\small 0.61}  & {\small 42.7}& {\small 25.8} & {\small 262} & {\small 156}  \\
\vspace{0.5 mm} {\small P3} & {\small 0.82} & {\small 0.67} & {\small 58 } &{\small 0.59} & {\small 35.5}  & {\small 22.0} & {\small 275} &{\small 178}   \\
\vspace{0.5 mm} {\small P4} & {\small 0.48} & {\small 0.51} & {\small 63} &{\small 0.81} & {\small 34.9}  & {\small 24.1} &  {\small 275}& {\small 207}  \\
\vspace{0.5 mm} {\small P5} & {\small 0.64} & {\small 0.58} & {\small 61} &{\small 0.84} & {\small 30.9}& {\small 23.1}  &  {\small 285} & {\small 155}   \\
\vspace{0.5 mm} {\small P6} & {\small 0.48} & {\small 0.55} & {\small 69} &{\small 1.04}  & {\small 26.1}& {\small 20.0} &  {\small 362}& {\small 272}  \\
\vspace{0.5 mm} {\small P7} & {\small 0.73} & {\small 0.67} & {\small 58} &{\small 1.02} & {\small 34.2} & {\small 26.7} &  {\small 265}&{\small 178} \\
\vspace{0.5 mm} {\small P8} & {\small 0.39} & {\small 0.47} & {\small 52} &{\small 1.01}  & {\small 39.2}& {\small 19.5} &  {\small 516}& {\small 369}  \\
\vspace{0.5 mm} {\small P9} & {\small 0.56} & {\small 0.59} & {\small 54} &{\small 0.83} & {\small 27.3} & {\small 32.1} &  {\small 293}&{\small 178} \\
\vspace{0.5 mm} {\small P10} & {\small 0.87} & {\small 0.63} & {\small 59} &{\small 0.42}  & {\small 43.7}& {\small 25.7} &  {\small 150}  & {\small 142}  \\ 
\end{tabular}}
\label{indicesP}
\end{table}  
\begin{table}[h!]
\caption{{\small Gait Indices for Control Subjects}}
\vspace{3 mm} 
\centering 
\resizebox{0.9\textwidth}{!}{  
\begin{tabular}{c c c c c c c c c c c}
\hline\hline
Control & $\mathbf{V}_{\rm n}$[s$^{-1}$] & $\mathbf{L}_{\rm n}$ & $\mathbf{S}$[\%] & $\mathbf{W}$[m] &  $\boldsymbol{\alpha}_K$[deg] & $\boldsymbol{\alpha}_H$[deg] & $\mathbf{D}_\mathrm{K}$[deg]& $\mathbf{D}_\mathrm{H}$[deg] \\[0.5ex]  
\hline 
\vspace{0.5 mm} {\small C1} & {\small 1.35} & {\small 0.78} & {\small 45} & {\small 0.64} & {\small 42.9} &  {\small 19.3} &  {\small 161}&{\small 155}  \\
\vspace{0.5 mm} {\small C2} & {\small 1.12} & {\small 0.78} & {\small 52} & {\small 0.54}  & {\small 50.1}& {\small 32.8} &  {\small 168}& {\small 124} \\
\vspace{0.5 mm} {\small C3} & {\small 1.12} & {\small 0.73} & {\small 53} &{\small 0.95}  & {\small 33.8}& {\small 26.9} &  {\small 206}& {\small 123} \\
\vspace{0.5 mm} {\small C4} & {\small 0.97} & {\small 0.66} & {\small 53} &{\small 0.79}  & {\small 37.8}& {\small 26.2} &  {\small 323}& {\small 223}  \\
\vspace{0.5 mm} {\small C5} & {\small 1.38} & {\small 0.90} & {\small 48} &{\small 0.41} & {\small 50.5} & {\small 32.8} &  {\small 160} & {\small 114} \\
\vspace{0.5 mm} {\small C6} & {\small 1.25} & {\small 0.81} & {\small 49} &{\small 0.65}  & {\small 42.6} & {\small 24.2}&  {\small 145}& {\small 117} \\
\vspace{0.5 mm} {\small C7} & {\small 1.19 } & {\small 0.80 } & {\small 48 } &{\small 0.58} & {\small 52.7}& {\small 30.6} &  {\small 164}& {\small 123}  \\
\vspace{0.5 mm} {\small C8} & {\small 1.21} & {\small 0.84} & {\small 46} &{\small 0.50}  & {\small 46.6}& {\small 31.1} &  {\small 156}& {\small 114}  \\
\vspace{0.5 mm} {\small C9} & {\small 1.16} & {\small 0.71} & {\small 54} &{\small 0.31}  & {\small 43.5}& {\small 27.2} &  {\small 195}& {\small 140} \\
\vspace{0.5 mm} {\small C10} & {\small 1.42} & {\small 0.77} & {\small 54 } &{\small 0.62}  & {\small 50.5} & {\small 38.1}&  {\small 170}& {\small 147} \\
\end{tabular}}
\label{indicesN}
\end{table} 

Next, we investigate the reliability of each index using the \textit{intraclass correlation coefficient} (ICC). Specifically, the ICC is calculated for both the patient and control groups. The results are given in Tables \ref{corr-Distance} and \ref{corr-Angles}. As it can be seen in both groups all indices have acceptable intraclass correlations with the exception of step width in MS patients. Also, the knee range of motion index possess a wide confidence interval. In order to statistically compare the differences between the value of indices corresponding to the patient and control groups, the paired $t$-test was used (see Tables \ref{corr-Distance} and \ref{corr-Angles}). All indices show statistically significant difference between the patient and control groups as signified by small $p$-values ($p\leq 0.05$).    
\begin{table}[h!]
\caption{{\small Statistics for time-distance-derived indices}}
\vspace{3 mm} 
\centering 
\resizebox{0.9\textwidth}{!}{  
\begin{tabular}{c c c c c}
\hline\hline
& {$\scriptsize\mathbf{V}_n$\scriptsize[s$^{-1}]$} & {\scriptsize$\mathbf{L}_n$} & {\scriptsize$\mathbf{S}\scriptsize$[\%]} & {\scriptsize$\mathbf{W}$[m]} \\[0.5ex] 
\hline 
  {\small MS Mean (SD)} & {\small 0.4 (0.14) } & {\small 0.6 (0.1)} & {\small 0.6 (0.05) } & {\small 0.8 (0.2)} \\

 {\small Control Mean (SD)} & {\small  1.2 (0.14)  } & {\small  0.8 (0.07) } & {\small  0.5 (0.03)  } & {\small  0.6 (0.18) } \\

{\small MS ICC (95\% CI)} & {\small  0.99 (0.984,0.998) } & {\small  0.93 (0.83,98)  } & {\small  0.71 (0.25,0.92) } & {\small  0.50 (0.63,0.96) } \\

{\small Control ICC (95\% CI)} & {\small  0.93 (0.84,0.98) } & {\small  0.90 (0.76,0.97) } & {\small  0.75 (0.38,0.93) } & {\small  0.80 (0.52,0.95) } \\

{\small Correlation Ambulation Score (95\% CI)} & {\small  -0.69  }{\small (-0.81,-0.50) } & {\small  0.54 } {\small  (0.30,0.71) } & {\small  -0.14 } {\small (-0.40,0.15) } & {\small  0.43 } {\small (0.17,0.64) } \\
{\small Correlation with MSWS (95\% CI)} & {\small  -0.86  }{\small (-0.91,-0.76) } & {\small  0.69  }{\small (0.50,81) }  & {\small  0.04  }{\small (-0.24,0.33) } & {\small  0.51  } {\small (0.26,0.70) } \\
{\small $t$-test $p$-value} & {\small $10^{-6}$ } & {\small $10^{-4}$ } & {\small $10^{-4}$ } &{\small $0.044$ } \\
\label{corr-Distance}
\end{tabular}}
\end{table}
\begin{table}[h!]
\caption{{\small Statistics for angle-derived indices}}
\centering 
\resizebox{0.9\textwidth}{!}{  
\vspace{3 mm} 
\begin{tabular}{c c c c c}
\hline\hline
& {\scriptsize$\boldsymbol{\alpha}_K\scriptsize$[deg]} & {\scriptsize$\boldsymbol{\alpha}_H\scriptsize$[\scriptsize deg]}   &{\scriptsize$\mathbf{D}_{\mathrm{K}}$[\scriptsize deg]} & {\scriptsize$\mathbf{D}_{\mathrm{H}}\scriptsize$[deg]} \\[0.5ex] 
\hline 
  {\small MS Mean (SD)}  &{\small 36 (5)} & {\small 23 (3)} & {\small 236 (68)} & {\small 210 (47) }\\

 {\small Control Mean (SD)}  &{\small 47 (9) } & {\small  30 (4) } & {\small  191 (54) } & {\small 156 (36) } \\

{\small MS ICC (95\% CI)} &{\small 0.61 (-0.01,0.90) } & {\small  0.92 (0.80,0.98)} & {\small  0.88 (0.69,0.97)} & {\small 0.78 (0.43,0.94)} \\

{\small Control ICC (95\% CI)}  & {\small 0.89 (0.74,0.97)} & {\small  0.98 (0.94,0.99)} & {\small  0.93 (0.82,0.97)} & {\small 0.82 (0.54,0.94)} \\

{\small Correlation Ambulation Score (95\% CI)}  &{\small -0.42 } {\small (-0.64,-0.14) } & {\small  -0.63 } {\small (-0.78,-0.41) } & {\small  0.66 } {\small (0.44,0.80) } & {\small 0.66 }{\small (0.45,0.80) }\\
{\small Correlation with MSWS (95\% CI)} & {\small -0.50 } {\small (-0.70,-0.24) } & {\small  -0.42 } {\small (-0.64,-0.14) } & {\small  0.64 } {\small (0.41,0.79) }  & {\small 0.62 }{\small (0.39,0.77) }\\
{\small $t$-test $p$-value} & {\small$0.0229$} & {\small $0.0067$} & {\small $0.0142$} & {\small$0.0026$}\\
\label{corr-Angles}
\end{tabular}}
\end{table}



Finally, we note that all considered indices with the exception of stance percentage, and hip range of motion show a high degree of correlation with the clinical ambulation score and MSWS. The stride width shows an acceptable degree of correlation with MSWS but not with the ambulation score. This observations signify the fact that the Kinect sensor can provide an objective framework for assessing gait abnormality that can be of clinical utility. The data can provide invaluable insight on the progression of the disease and response to treatment.
\clearpage
\bibliographystyle{IEEEtran}
\bibliography{MS}
\end{document}

%% file: title.tex
\begin{center}
\thispagestyle{empty} \baselineskip 22pt { {\huge\bf Gait Assessment for Multiple Sclerosis Patients Using Microsoft Kinect}
\ \\ \ \\
\vfill
{\Large\bf Farnood Gholami$^1$\\ Daria A. Trojan, MD, MSc$^2$\\ J\'{o}zsef K\"{o}vecses, PhD$^1$\\ Wassim M. Haddad, PhD$^3$\\ Behnood Gholami, PhD$^3$\\}
\ \\ 
{\large $^1$Department of Mechanical Engineering and Centre for Intelligent Machines, McGill University, Montreal, QC H3A 2K6, Canada\\
$^2$Department of Neurology and Neurosurgery, Montreal Neurological Institute and Hospital, McGill University Health Centre, McGill University, Montreal, QC H3A 2B4, Canada\\
$^3$AreteX Engineering, New York, NY 10013, United States\\}
\vfill August 10, 2015\\} \vfill
\end{center}
\newpage

%% file: report_revised.bbl
\begin{thebibliography}{10}
\providecommand{\url}[1]{#1}
\csname url@samestyle\endcsname
\providecommand{\newblock}{\relax}
\providecommand{\bibinfo}[2]{#2}
\providecommand{\BIBentrySTDinterwordspacing}{\spaceskip=0pt\relax}
\providecommand{\BIBentryALTinterwordstretchfactor}{4}
\providecommand{\BIBentryALTinterwordspacing}{\spaceskip=\fontdimen2\font plus
\BIBentryALTinterwordstretchfactor\fontdimen3\font minus
  \fontdimen4\font\relax}
\providecommand{\BIBforeignlanguage}[2]{{%
\expandafter\ifx\csname l@#1\endcsname\relax
\typeout{** WARNING: IEEEtran.bst: No hyphenation pattern has been}%
\typeout{** loaded for the language `#1'. Using the pattern for}%
\typeout{** the default language instead.}%
\else
\language=\csname l@#1\endcsname
\fi
#2}}
\providecommand{\BIBdecl}{\relax}
\BIBdecl

\bibitem{Hobart}
J.~C. Hobart, A.~Riazi, D.~L. Lamping, R.~Fitzpatrick, and A.~J. Thompson,
  ``Measuring the impact of ms on walking ability: the 12-item ms walking scale
  {(MSWS-12)},'' \emph{Neurology}, vol.~60, pp. 31--36, 2003.

\bibitem{Kurtzke}
J.~F. Kurtzke, ``Rating neurologic impairment in multiple sclerosis,''
  \emph{Neurology}, vol.~33, no.~11, p. 1444, 1983.

\bibitem{Holden}
M.~K. Holden, K.~M. Gill, and M.~R. Magliozzi, ``Gait assessment for
  neurologically impaired patients,'' \emph{Phys. Therapy}, vol.~66, pp.
  1530--1539, 1986.

\bibitem{Remelius}
J.~G. Remelius, J.~Hamill, J.~Kent-Braun, and R.~E.~A. Van~Emmerik, ``Gait
  initiation in multiple sclerosis,'' \emph{Motor Control}, vol.~12, no.~2, pp.
  93--108, 2008.

\bibitem{Sacco}
R.~Sacco, R.~Bussman, P.~Oesch, J.~Kesselring, and S.~Beer, ``Assessment of
  gait parameters and fatigue in ms patients during inpatient rehabilitation:
  {A} pilot trial,'' \emph{J. Neurology}, vol. 258, no.~5, pp. 889--894, 2011.

\bibitem{Givon}
U.~Givon, G.~Zeilig, and A.~Achiron, ``Gait analysis in multiple sclerosis:
  Characterization of temporal–spatial parameters using gaitrite functional
  ambulation system,'' \emph{Gait \& Posture}, vol.~29, pp. 138--142, 2009.

\bibitem{Gehlsen}
G.~Gehlsen, K.~Beekman, N.~Assmann, D.~Winant, M.~Seidle, and A.~Carter, ``Gait
  characteristics in multiple sclerosis: {P}rogressive changes and effects of
  exercise on parameters,'' \emph{Arch. Phys. Med. Rehab.}, vol.~67, pp.
  536--539, 1986.

\bibitem{Benedetti}
M.~G. Benedetti, R.~Piperno, L.~Simoncini, P.~Bonato, A.~Tonini, and
  S.~Giannini, ``Gait abnormalities in minimally impaired multiple sclerosis
  patients,'' \emph{Mult. Sclerosis J.}, vol.~5, no.~5, pp. 363--368, 1999.

\bibitem{Huisinga}
J.~M. Huisinga, K.~K. Schmid, M.~Filipi, and S.~N., ``Gait mechanics are
  different between healthy controls and patients with multiple sclerosis,''
  \emph{J. Appl. Biomech.}, vol.~29, pp. 303--311, 2013.

\bibitem{Kelleher}
K.~J. Kelleher, W.~Spence, S.~Solomonidis, and D.~Apatsidis, ``The
  characterisation of gait patterns of people with multiple sclerosis,''
  \emph{Disab. Rehab.}, vol.~32, no.~15, pp. 1242–--1250, 2010.

\bibitem{Shotton}
J.~Shotton, T.~Sharp, A.~Kipman, M.~Fitzgibbon, A.~Finocchio, A.~Blake, and
  M.~Cook, ``Real-time human pose recognition in parts from a single depth
  image,'' in \emph{IEEE Conf. Comp. Vis. Patt. Recog.}, 2011, pp. 1297--1304.

\bibitem{Baena}
A.~Ferna\'ndez-Baena, A.~Susin, and X.~Lligadas, ``Biomechanical validation of
  upper-body and lower-body joint movements of kinect motion capture data for
  rehabilitation treatments,'' \emph{Proc. Int. Conf. Intell. Networking
  Collab. Syst.}, pp. 656--661, 2012.

\bibitem{Clark}
R.~A. Clark, Y.~Pua, K.~Fortin, C.~Ritchie, K.~E. Webster, L.~Denehy, and A.~L.
  Bryant, ``Validity of the microsoft kinect for assessment of postural
  control,'' \emph{Gait \& Posture}, vol.~36, pp. 372--377, 2012.

\bibitem{Galna}
B.~Galna, G.~Barry, D.~Jackson, D.~Mhiripiri, P.~Olivier, and L.~Rochester,
  ``Accuracy of the microsoft kinect sensor for measuring movement in people
  with parkinson's disease,'' \emph{Gait \& Posture}, vol.~39, pp. 1062--1068,
  2014.

\bibitem{Pfueller}
C.~Pf{\"{u}}eller, K.~Otte, S.~Mansow-Model, F.~Paul, and A.~Brandt,
  ``Kinect-based analysis of posture, gait and coordination in multiple
  sclerosis patients,'' \emph{Neurology}, vol.~80, 2013.

\bibitem{Souza}
M.~D. Souza, C.~Kamm, J.~Burggraaff, P.~Tewarie, B.~Glocker, J.~Dorn, T.~Vogel,
  C.~Morrison, A.~Sellen, M.~Machacek, P.~Chin, B.~Uitdehaag, A.~Criminisi,
  F.~Dahlke, C.~Polman, and L.~Kappos, ``Assessment of disability in multiple
  sclerosis using the {K}inect-camera system: {A} proof-of-concept study,''
  \emph{Neurology}, vol.~82, no.~10, p. 139, 2014.

\bibitem{Behrens2}
B.~J., C.~Pf{\"{u}}ller, S.~Mansow-Model, K.~Otte, F.~Paul, and A.~U. Brandt,
  ``Using perceptive computing in multiple sclerosis - the short maximum speed
  walk test,'' \emph{J. Neuroeng. Rehab.}, vol.~11, no.~1, p.~89, 2014.

\bibitem{Kontschieder}
P.~Kontschieder, J.~F. Dorn, C.~Morrison, R.~Corish, D.~Zikic, A.~Sellen,
  M.~D'Souza, C.~P. Kamm, J.~Burggraaff, P.~Tewarie, T.~Vogel, M.~Azzarito,
  B.~Glocker, P.~Chin, F.~Dahlke, C.~Polman, L.~Kappos, B.~Uitdehaag, and
  A.~Criminisi, ``Quantifying progression of multiple sclerosis via
  classification of depth videos,'' \emph{Med. Image Comput. Comput. Assist.
  Interv. (MICCAI)}, vol.~17, no.~2, pp. 429--437, 2014.

\bibitem{Muller}
M.~M{\"{u}}ller, ``Dynamic time warping,'' in \emph{Information Retrieval for
  Music and Motion}, 2007, pp. 69--84.

\bibitem{Rabiner}
L.~R. Rabiner and B.~H. Juang, \emph{Fundamentals of Speech Recognition}.\hskip
  1em plus 0.5em minus 0.4em\relax Prentice Hall Signal Processing Series,
  1993.

\bibitem{Veeraraghavan}
A.~Veeraraghavan, A.~K. Roy-Chowdhury, and R.~Chellappa, ``Matching shape
  sequences in video with applications in human movement analysis,'' \emph{IEEE
  Trans. Patt. Anal. Mach. Intell.}, vol.~27, no.~12, pp. 1896--1909, 2005.

\bibitem{Blackburn}
J.~Blackburn and E.~Ribeiro, ``Human motion recognition using isomap and
  dynamic time warping,'' in \emph{Human Motion: Understanding, Modeling,
  Capture and Animation, Lecture Notes in Computer Science}, Rio de Janeiro,
  Brazil, 2007, vol.~27, no. 4814, pp. 285--298.

\end{thebibliography}
